\documentclass[11pt]{article}

\usepackage[final]{acl}

\usepackage{times}
\usepackage{latexsym}

\usepackage[T1]{fontenc}

\usepackage[utf8]{inputenc}

\usepackage{microtype}

\usepackage{inconsolata}

\usepackage{graphicx}
\usepackage{amssymb}
\usepackage{amsmath}
\usepackage{svg}
\usepackage{tabularx}
\usepackage{booktabs}
\usepackage{enumitem}
\usepackage{multirow}
\usepackage{hyperref}

\usepackage{subcaption}
\usepackage{pgfplots}
\pgfplotsset{compat=1.18}

\usepackage[acronym,nolist]{glossaries}
\glsdisablehyper
\newacronym{dvp}{DVP}{Deep Visual Processor}
\newacronym{vrp}{VRP}{Visual Relevance Predictor}
\newacronym{kd}{KD}{Knowledge Distillation}
\newacronym[
  plural={Determinantal Point Processes},
  shortplural={DPPs}
]{dpp}{DPP}{Determinantal Point Process}

\title{Towards Visually Grounded Multimodal Summarization via Cross-Modal Transformer and Gated Attention}

\author{
  Abid Ali\thanks{Corresponding Author} \and
  Diego Moll{\'a}-Aliod \and
  Usman Naseem \\
  School of Computing, Macquarie University, Sydney, Australia \\
  \texttt{abidmeeraj@gmail.com, diego.molla-aliod@mq.edu.au, usman.naseem@mq.edu.au}
}

\begin{document}
\maketitle
\begin{abstract}

Multimodal summarization requires models to jointly understand textual and visual inputs to generate concise, semantically coherent summaries. Existing methods often inject shallow visual features into deep language models, leading to representational mismatches and weak cross-modal grounding. We propose a unified framework that jointly performs text summarization and representative image selection. Our system, \texttt{SPeCTrA-Sum} (Sampler Perceiver with Cross-modal Transformer and gated Attention for Summarization), introduces two key innovations. First, a \gls{dvp} aligns the visual encoder with the language model at corresponding depths, enabling hierarchical, layer-wise fusion that preserves semantic consistency. Second, a lightweight \glsentrylong{vrp} (\glsentryshort{vrp}) selects salient and diverse images by distilling soft labels from a \glsentrylong{dpp} (\glsentryshort{dpp}) teacher. \texttt{SPeCTrA-Sum} is trained using a multi-objective loss that combines autoregressive summarization, cross-modal alignment, and \glsentryshort{dpp}-based distillation. Experiments show that our system produces more accurate, visually grounded summaries and selects more representative images, demonstrating the benefits of depth-aware fusion and principled image selection for multimodal summarization.

\end{abstract}

\section{Introduction}

\begin{figure}[!t]
\vspace{-0.2cm}
    \centering
    \includegraphics[width=0.48\textwidth]{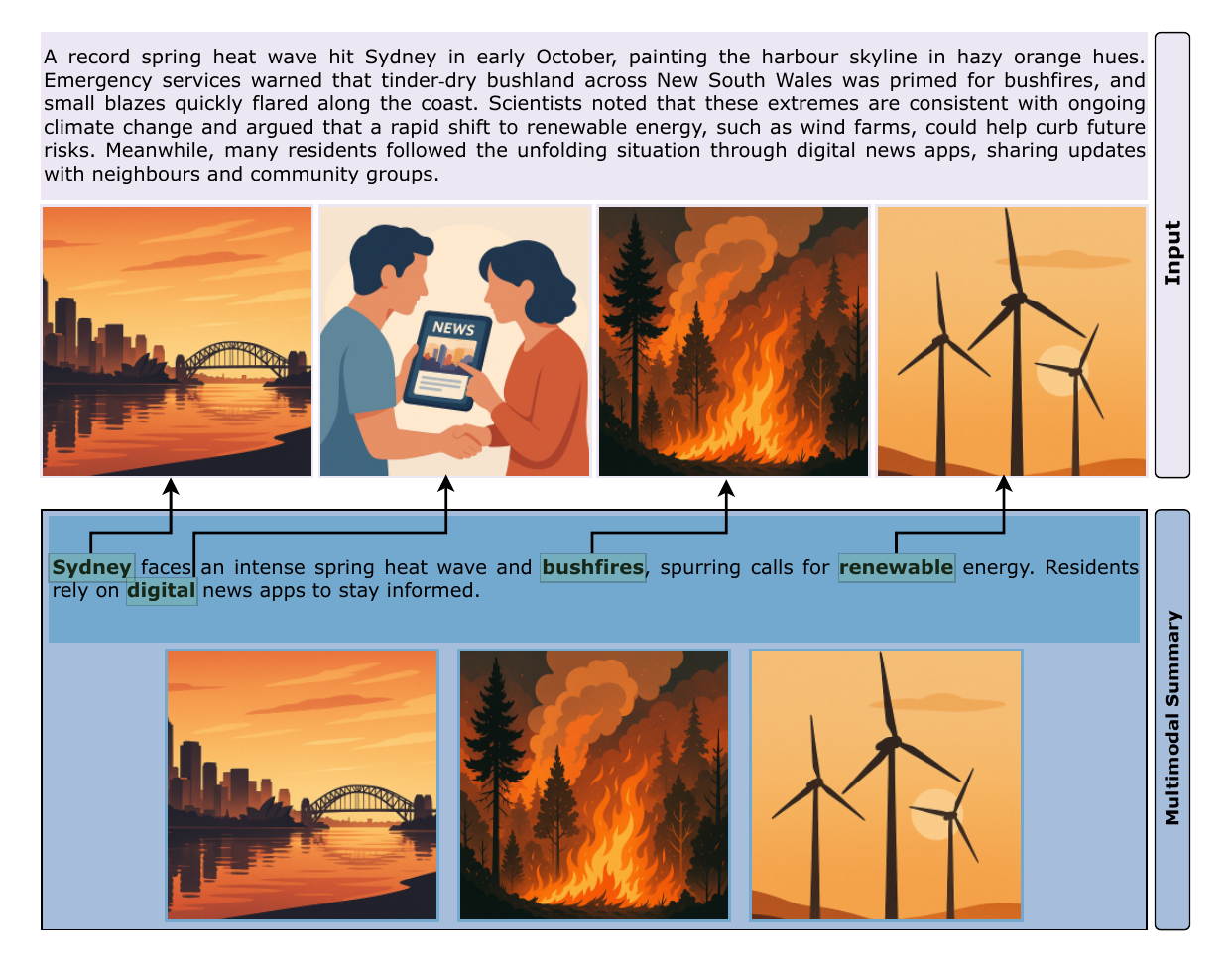}
    \vspace{-0.3cm}
    \caption{An Illustration of a Multimodal System with Multimodal Output.}
    \label{fig:MSMO_Example}
    \vspace{-0.4cm}
\end{figure}

In today's information-rich environment, users are often overwhelmed by the need to quickly process large volumes of multimodal content. This includes news articles with embedded images, blog posts featuring photo galleries, or technical reports combining diagrams and captions. While traditional summarization models focus exclusively on textual compression, real-world documents often span both visual and textual modalities \cite{Zhu2018MSMO:Output}. As a result, multimodal summarization, the task of generating concise summaries conditioned on both text and images, has emerged as an increasingly important research direction. Figure~\ref{fig:MSMO_Example} shows an example of multimodal summarization with multimodal output.

Recent surveys emphasize that multimodal summarization must effectively integrate heterogeneous inputs from different modalities \cite{Atharva2023TheSummarization, Zhang2023MultimodalSurvey, Jangra2023ASummarization}. However, despite substantial progress in multimodal learning, many models continue to exhibit fundamental limitations in cross-modal integration, largely due to a persistent semantic gap between visual and textual representations \cite{Zhu2018MSMO:Output}. 

In practice, visual features are often only shallowly fused into deep language models, resulting in weak semantic alignment between modalities \cite{Liu2025Symmetry-AwareEvaluation}. Consequently, image representations frequently fail to capture deeper textual abstractions \cite{Chen2025MultimodalShallower}, limiting effective cross-modal grounding. For example, models may struggle to reliably associate visual content with its corresponding textual descriptions \cite{Li2025STI-Bench:Understanding}. As a result, vision and language remain only loosely coupled in many existing systems, underscoring the need for architectures that more effectively bridge this representational divide.

Another key challenge is selecting images that meaningfully contribute to the summary. Source documents often include many images, some of which are redundant or peripheral. Thus, identifying a representative, non-redundant subset is essential. This involves a dual goal: selecting images that are both individually relevant and collectively diverse. In related areas like image summarization and keyframe extraction, \glspl{dpp} have proven effective in modeling this relevance--diversity trade-off \cite{Celis2018FairSummarization, Cho2019Multi-DocumentRepresentations}.

Motivated by the inherent coupling between cross-modal alignment and visual subset selection, we propose a unified architecture that jointly optimizes abstractive summary generation and the selection of image subsets that best support the generated text. Our key contributions are:

\begin{itemize}[noitemsep,leftmargin=*]
    \item \textbf{Joint multimodal summarization and image selection:} We model image selection as an integral component of the summarization process rather than a post-hoc step. Specifically, we introduce the \gls{vrp}, which jointly learns to generate a summary $Y$ and select a representative image subset $I^* \subseteq I$ from the input text $X$ and image set $I$, enabling tighter alignment between textual and visual outputs.

    \item \textbf{Knowledge-distilled image selector with DPP supervision:} To move beyond heuristic image scoring, we employ a \gls{dpp}-based teacher that jointly models text--image relevance and inter-image diversity to produce soft inclusion probabilities. These pseudo-labels are used to train the \gls{vrp}, enabling it to efficiently select relevant and non-redundant image subsets at inference time.

    \item \textbf{Multi-objective training for cross-modal alignment and diversity-aware selection:} We train the model with a unified objective that combines (i) an autoregressive summarization loss, (ii) a cross-modal alignment loss between generated text and visual features, and (iii) a distillation loss derived from the \gls{dpp} teacher. Together, these objectives encourage the model to jointly optimize textual quality, visual grounding, and diversity-aware image selection.
    
\end{itemize}

We evaluate our method on the MSMO dataset \cite{Zhu2018MSMO:Output} and demonstrate that it (i) improves standard summarization metrics over baselines without image selection, (ii) produces more visually grounded summaries with higher image--text relevance, and (iii) selects image subsets that better balance relevance and diversity compared to heuristic or text-only approaches.

\section{Related Work}\label{sec:related_work}
\subsection{Text and Multimodal Summarization}
Early abstractive summarization models framed the summarization task as sequence modeling with attention. \citet{Rush2015ASummarization} introduced a neural attention-based model for sentence-level summarization, outperforming extractive baselines. Follow-ups \cite[for example]{Nallapati2016AbstractiveBeyond,See2017GetNetworks} improved factuality and coverage using pointer-generator mechanisms and copy-aware decoding.

Multimodal summarization extends this task to include visual inputs such as images or videos. \citet{Zhu2018MSMO:Output} formally defined the multimodal setting and proposed a joint model for text generation and image selection, along with the MSMO dataset. Later works explored multimodal attention to improve informativeness and grounding, showing that integrating image features via attention can enhance summary quality \cite{Chen2018AbstractiveRNN, Li2018Multi-modalFiltering}.

\subsection{Multimodal Fusion and Alignment in Vision-Language Models}
Early Vision-Language (VL) models like ViLBERT \cite{Lu2019ViLBERT:Tasks} and LXMERT \cite{Tan2019LXMert:Transformers} used dual-stream encoders with cross-attention to learn joint multimodal representations, improving performance on tasks like VQA. More recent models, such as Flamingo \cite{Alayrac2022Flamingo:Learning}, integrate pretrained vision and language modules using gated cross-attention, enabling few-shot generation from interleaved image--text inputs.

Several studies \cite{Cao2020BehindModels, Bugliarello2021MultimodalBERTs} show that deeper layers tend to prioritize textual representations while paying limited attention to visual inputs, motivating the need for more effective layer-aligned fusion strategies.

\subsection{Image Selection and Visual Importance}
Multimodal summarization requires identifying which images meaningfully support the generated summary. Including all available images can dilute relevance or introduce redundancy, motivating the need for selective conditioning. \citet{Liang2023Summary-OrientedSummarization} improve visual encoder utility via auxiliary objectives like vision-to-summary prediction and masked image modeling. CFSum \cite{Xiao2023CFSum:Summarization} filters uninformative images through a pre-selection module and modulates attention with visual complement units. DIUSum \cite{Xiao2024DIUSum:Summarization} further predicts image usefulness and dynamically gates visual reliance during decoding. Together, these approaches highlight the benefits of targeted visual integration for content quality and efficiency.

\subsection{Knowledge Distillation and Multi-Objective Learning}
\gls{kd} \cite{Hinton2015DistillingNetwork} trains compact models to mimic larger ones via softened output distributions. In multimodal settings, \gls{kd} has enabled cross-modal supervision; for instance, \citet{Gupta2016CrossTransfer} transferred knowledge from labeled RGB to depth/optical flow streams using paired data.

For VL pretraining, ALBEF \cite{Li2021AlignDistillation} combined contrastive alignment with momentum-based self-distillation to learn from noisy web pairs. More recent work explores distilling large CLIP models into smaller ones (e.g., CLIP-KD \cite{Yang2024Clip-kd:Distillation, Pei2023Clipping:Retrieval}), extending \gls{kd} to video--language tasks.

\noindent In summary, prior work has advanced multimodal summarization through improved fusion strategies, image filtering, and cross-modal alignment. However, several challenges remain. Image selection is often heuristic or treated as a separate stage from text generation, limiting joint optimization and alignment. Moreover, most approaches prioritize relevance while overlooking diversity within the selected image set, resulting in redundant or less informative visual summaries. To address these limitations, we propose a unified framework that integrates gated, depth-aware fusion, distillation from a diversity-aware teacher, and multi-objective training, enabling the generation of coherent summaries supported by informative and complementary visual content.

\section{Methodology}\label{sec:methodology}

\begin{figure*}[!t]
    \centering
    \includegraphics[width=0.9\textwidth]{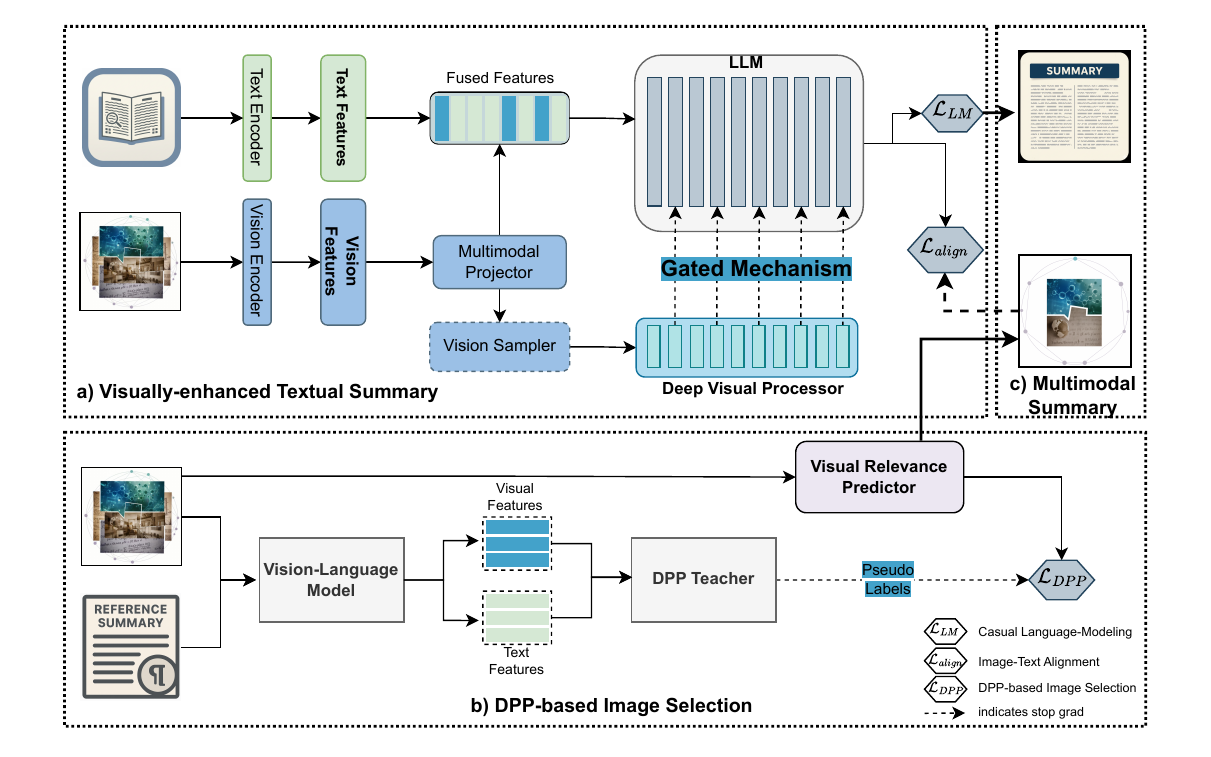}
    \vspace{-0.35cm}
    \caption{An overview of SPeCTrA-Sum Framework.}
    \vspace{-0.35cm}
    \label{fig:framework}
\end{figure*}

As shown in Figure~\ref{fig:framework}, \texttt{SPeCTrA-Sum} consists of two main stages: (a) representation and summarization, where the input text and images are encoded and fused to produce a summary conditioned on both modalities; and (b) image selection and alignment, where a subset of salient images is selected and aligned with the generated summary. The entire system is trained end-to-end using a multi-objective loss that jointly optimizes for summary quality, visual representativeness, and cross-modal semantic alignment.

\subsection{Problem Setup and Base Architecture}

A multimodal document consists of text $X = \{x_1, \dots, x_{T_x}\}$ and images $\mathcal{I} = \{I_1, \dots, I_M\}$. The objective is to generate an abstractive summary $Y = \{y_1, \dots, y_{T_y}\}$ and a representative image subset $\mathcal{I}^\star \subseteq \mathcal{I}$ that reflects salient visual content aligned with the summary. Formally, the model learns a conditional mapping:

\vspace{-0.3cm}
\begin{equation}
\label{eq:problem_def}
f_\theta: (X,\,\mathcal{I}) \;\longrightarrow\; (Y,\,\mathcal{I}^\star),
\end{equation}
\vspace{-0.3cm}

where $Y = f_\theta^{\text{text}}(X, \mathcal{I})$ is produced by the decoder, and $\mathcal{I}^\star = f_\theta^{\text{vis}}(\mathcal{I})$ is predicted by the \gls{vrp}. Both outputs are trained jointly to ensure that $\mathcal{I}^\star$ is (i) semantically aligned with $Y$ and (ii) diverse, promoting textual--visual coherence and visual complementarity.

\paragraph{Backbone Architecture.}
We adopt LLaVA-OneVision \cite{Li2025LLaVA-OneVision:Transfer} as our multimodal scaffold, which uses Qwen-2 \cite{Wang2024Qwen2-VL:Resolution} as the causal language model (LLM) for generation. In this setup, a frozen vision encoder (SigLIP by \citet{Zhai2023SigmoidPre-training}) is connected to the LLM via a multimodal projector, mapping visual features into the same token embedding space. This enables token-level concatenation of image and text inputs, a design widely used in LLaVA and MiniGPT-4 \cite{Liu2023VisualTuning, Zhu2024MINIGPT-4:MODELS}.

Each visual vector is then projected into the LLM's token space via a learned linear mapping:

\vspace{-0.3cm}
\begin{equation}
\tilde{\mathbf{v}}^{(m)}_t = P\left(\mathbf{v}^{(m)}_t\right).
\label{eq:projector}
\end{equation}

After projection, features from both modalities are concatenated. This ``projection-then-concatenation'' strategy aligns vision and language tokens in a shared space. 

\paragraph{Limitations of Pure Concatenation.}
Concatenating all projected visual tokens $\tilde{v}$ at the prefix of a decoder-only LLM introduces two major issues. First, in causal architectures, prefix tokens exert less influence in deeper layers due to attention bias toward recent tokens. Second, many visual tokens may be irrelevant or redundant, resulting in inefficient attention usage. Moreover, the raw projector outputs often exhibit a representation gap relative to the deeply transformed LLM hidden states, leading to poor cross-modal alignment.

These limitations motivate our proposed approach, which introduces hierarchically aligned and semantically filtered fusion mechanisms via the \glsentrylong{dvp} (\glsentryshort{dvp}) and \gls{vrp}.

\subsection{Proposed Method}\label{sec:proposed_method}
We propose a unified framework that enhances multimodal summarization through five key components. Three modules are introduced to improve text--image fusion and representation alignment: (i) a vision sampler that filters out uninformative visual inputs, (ii) a DVP that enables layer-wise alignment between visual and textual features, and (iii) a Gated Integration mechanism that allows the model to dynamically control the contribution of visual features at different decoding depths. To further enable output-level image selection, we incorporate two additional components: (iv) a \gls{dpp}-based Pseudo-Label Generator that produces soft inclusion probabilities encouraging both relevance and diversity, and (v) a lightweight image selector trained to learn from these labels, enabling efficient selection of salient, non-redundant images at inference time.

\subsubsection{Vision Sampler}
To reduce redundancy and retain only the most informative visual signals, we compress each image's patch grid into a fixed set of latent tokens using a Perceiver-style cross-attention bottleneck. Unlike top-$K$ patch selection, this mechanism allows the model to learn what to retain via trainable latent queries.

Our design follows the Perceiver family of models \cite{Jaegle2022PERCEIVEROUTPUTS}, where learned latent vectors attend to the full input via cross-attention and are refined through self-attention. Flamingo \cite{Alayrac2022Flamingo:Learning} similarly uses a Perceiver Resampler to compress visual inputs into a fixed set of media tokens for efficient language integration.

Formally, for an image $I_m$ represented by patch features $\mathbf{V}^{(m)}\!\in\!\mathbb{R}^{T_v\times d_v}$, and corresponding positional encodings $\mathbf{P}^{(m)}\!\in\!\mathbb{R}^{T_v\times d_v}$, we produce a compressed representation using a Perceiver stack of depth $D$. The vision sampler maintains a trainable latent array $\mathbf{L}^{(0)}\!\in\!\mathbb{R}^{L\times d}$, where $L$ is the number of compressed visual tokens output per image. These latents serve as compact, semantically rich visual summaries, passed downstream to the language model via multimodal fusion.

\subsubsection{Deep Visual Processor}
To address the representation gap between shallow visual embeddings and deep language model activations, we introduce a \gls{dvp} that refines visual features through a stack of transformer layers aligned with the LLM's depth.

Specifically, the compressed visual tokens produced by the Vision Sampler are processed through a sequence of $L$ transformer blocks:

\vspace{-0.5cm}
\begin{equation}
\hat{\mathbf{v}}^{(\ell)}=\mathrm{DVP}^{(\ell)}\!\left(\hat{\mathbf{v}}^{(\ell-1)}\right),\quad \ell=1,\dots,L,
\label{eq:dvp}
\end{equation}

where each \gls{dvp} block consists of a self-attention layer followed by a feed-forward MLP, and matches the hidden dimension $d$ of the corresponding LLM layer. This results in depth-aligned visual representations $\hat{\mathbf{v}}^{(\ell)}$, which evolve in parallel with the LLM's hidden states $\mathbf{v}^{(\ell)}$. This alignment enables layer-wise multimodal fusion, ensuring that visual information is semantically compatible with the abstraction level of each LLM layer.

\subsubsection{Layer-Aligned Gated Cross-Attention}

To enable controlled and depth-aware fusion of vision into the LLM, we insert gated cross-attention modules at every $n$-th layer in the decoder. At each injection point $\ell \in \{n, 2n, \dots\}$, depth-aligned visual tokens $\hat{\mathbf{v}}^{(\ell)}$ attend to the LLM hidden states via a tanh-gated residual connection, inspired by Flamingo's cross-attention strategy \cite{Alayrac2022Flamingo:Learning}.

Each gate is initialized near zero, allowing the model to initially preserve the behavior of the base LLM and gradually learn to integrate visual input. This mechanism supports progressive, learnable fusion aligned with the semantic hierarchy of the decoder. Full attention equations and gating details are provided in Appendix~\ref{sec:gated_cross_attention}.

\subsubsection{Visual Relevance Predictor (VRP)}
The \gls{vrp} is a lightweight module that selects a subset of images that are both semantically relevant and mutually diverse, enabling the summarizer to condition on the most informative visual content. \gls{vrp} is trained in a student--teacher distillation framework, where a principled \gls{dpp} \cite{Macchi1975TheProcesses, Kulesza2012DeterminantalLearning} acts as the teacher, producing soft selection probabilities that encode a trade-off between quality and diversity. A compact student network is trained to approximate these probabilities using only image embeddings, allowing for efficient, text-free inference that preserves the \gls{dpp}'s inductive biases.

\paragraph{DPP pseudo-labels.}
During training, the \gls{dpp} acts as a teacher that assigns each image a soft inclusion probability $\pi_i \in [0,1]$ based on its relevance to the summary and its redundancy with other images. These probabilities encode three key properties:

\begin{enumerate}[noitemsep,leftmargin=*]
    \item \textbf{Relevance}: Images closely aligned with the text receive higher probabilities.
    \item \textbf{Diversity}: Redundant images are down-weighted through \gls{dpp} structure.
    \item \textbf{Cardinality control}: The expected set size $\sum_i \pi_i \approx \mu$ is enforced explicitly.
\end{enumerate}

The \gls{dpp} kernel is constructed using image-text relevance scores and an RBF-based \cite{Lowe1988MultivariableNetworks} diversity matrix over image embeddings. We refer readers to Appendix~\ref{sec:dpp_derivation} for full derivation of the kernel and marginal inclusion computation.

\paragraph{Student Network.}
The student network, \gls{vrp}, is a small feed-forward predictor that maps each image embedding to a scalar selection logit. It operates independently on each image and does not require textual input at inference time. Given a normalized image embedding $v_i$, the predictor outputs:

\vspace{-0.4cm}
\begin{equation}
    z_i = f_{\mathrm{VRP}}(v_i),
\end{equation}

where $f_{\text{VRP}}$ is a two-layer MLP with GELU and dropout. The output $p_i$ approximates the teacher's inclusion score $\pi_i$ without requiring text. Since the \gls{dpp} already captures set-level interactions, the student is trained to mimic these marginals independently for each image (see Section~\ref{sec:loss_DRP} for the image selection loss).

\paragraph{Inference Efficiency.}
While \gls{dpp}-based selection requires expensive matrix operations and text conditioning ($\mathcal{O}(K^3)$, where $\mathcal{K}$ is the matrix dimension), the distilled \gls{vrp} enables text-free, constant-time selection per image. This achieves practical inference speed while retaining \gls{dpp}-style inductive biases, including relevance, diversity, and cardinality control. Empirically, the student achieves a strong approximation of the teacher and improves the quality of image subsets used for summarization.

\subsection{Training Objective}
The model is trained using a multi-objective (multimodal) loss that jointly optimizes three components:
\begin{enumerate}[noitemsep,leftmargin=*]
    \item an autoregressive language modeling loss for summary generation,
    \item an image--text alignment loss to encourage cross-modal coherence, and
    \item a distillation loss for training the \gls{vrp} to mimic the \gls{dpp} teacher.
\end{enumerate}

Formally, the loss is defined as:

\vspace{-0.4cm}
\begin{equation}
\begin{aligned}
\mathcal{L}_{\text{MM}}
&=
\underbrace{\mathcal{L}_{\text{LM}}}_{\text{causal LM}}
\;+\;
\lambda_{\text{align}}\,\underbrace{\mathcal{L}_{\text{align}}}_{\text{image--text alignment}}
\\
&\quad
+\;
\lambda_{\text{VRP}}\,\underbrace{\mathcal{L}_{\text{DPP}}}_{\text{image selection}} ,
\end{aligned}
\label{eq:total_loss}
\end{equation}

where $\lambda_{\text{align}}$ and $\lambda_{\text{VRP}}$ are hyperparameters that weight the auxiliary objectives relative to the main summarization loss.

\subsubsection{Autoregressive summarization}
Given a target token sequence $Y=\{y_1,\dots,y_T\}$, source text $x$, and a corresponding set of images $\mathcal{I}$, the model is trained using teacher forcing to maximize the likelihood of the correct summary tokens. The autoregressive language modeling loss is defined as:

\vspace{-0.8cm}
\begin{equation}
\label{eq:ar_loss}
\mathcal{L}_{\text{LM}}
\;=\;
-\sum_{t=1}^{T}\log p_\theta\!\left(y_t \,\middle|\, y_{<t},\, x,\, \{\mathcal{I}\}\right),
\end{equation}

where $p_\theta$ denotes the model's conditional probability distribution over tokens, given the multimodal input and previously generated tokens.

\subsubsection{Image-Text Alignment}
To ensure that generated summaries are semantically consistent with selected images, we introduce a global alignment objective between a projected decoder representation and a frozen visual embedding. Specifically, we align the decoder's mean-pooled hidden state to the average SigLIP embedding of the selected images, using a SigLIP-style contrastive loss.

The alignment encourages the decoder to ground textual output in visual semantics, while penalizing mismatches with unrelated image sets. Full formulation, including projection and contrastive loss computation, is provided in Appendix~\ref{sec:image_text_alignment_objective}.

\subsubsection{Image Selection via DPP Distillation}\label{sec:loss_DRP}
To learn image selection that is both relevant and diverse, we train the \gls{vrp} module to approximate soft inclusion probabilities generated by a \gls{dpp} teacher. These probabilities encode a principled balance between text-image relevance, visual redundancy, and set cardinality.

During training, we compute the teacher probabilities using a \gls{dpp} kernel constructed from text--image similarity and inter-image distances in a frozen SigLIP embedding space. The \gls{vrp} then learns to match these soft labels using a calibrated cross-entropy loss, along with a regularization term to enforce target subset size $\mu$.

This distillation allows the \gls{vrp} to internalize the \gls{dpp}'s inductive biases while enabling fast, text-agnostic image selection at inference. Full derivations and kernel construction are provided in Appendix~\ref{sec:dpp_image_selection}.

\section{Results}\label{sec:results}
\subsection{Implementation Details}
We trained our model using a batch size of 1 with the Adafactor optimizer, following \citet{Marek2025SmallWasteful}, who showed that smaller batch sizes can be more effective than larger ones in certain settings. Training was controlled by step count, where one epoch corresponds to approximately 295k steps. Different systems were trained up to 360k steps, and the best model was selected based on validation loss. Further technical details are provided in Appendix~\ref{app:technical_details} and Appendix~\ref{sec:overhead_analysis}.\footnote{Code: https://github.com/abidmeeraj/SPeCTrA-Sum}

\begin{table*}[!t]
\vspace{-0.2cm}
\centering
\small
\begin{tabular}{llccccc}
\toprule
\textbf{} & \textbf{Model} & \textbf{ROUGE-1} & \textbf{ROUGE-2} & \textbf{IP} & \textbf{MaxSim} & \textbf{MMAE} \\
\midrule
\multirow{10}{*}{\textbf{Baselines}}
  & ATG \shortcite{Zhu2018MSMO:Output}    & 40.63 & 18.12 & 59.28 & 25.82 & 3.35 \\
  & ATL \shortcite{Zhu2018MSMO:Output}   & 40.86 & 18.27 & 62.44 & 13.26 & 3.26 \\
  & HAN \shortcite{Zhu2018MSMO:Output}   & 40.82 & 18.30 & 61.83 & 12.22 & 3.25 \\
  & MOF \shortcite{Zhu2020MultimodalReference} & 41.20 & 18.33 & 65.45 & 26.38 & 3.37 \\
  & UniMS \shortcite{Zhang2022Unims:Distillation} & 42.94 & 20.50 & 69.38 & 29.72 & - \\
  & SITA \shortcite{Jiang2023ExploitingSummarization} &  43.64 & 20.53 & 76.41 & 33.47 & 3.37 \\
  & BART-VGG \shortcite{Cui2024AlignSummarization} & 43.75 & 20.70 & - & - & - \\
  & ViL-Sum \shortcite{Cui2024AlignSummarization} & 44.29 & 20.96 & 66.27 & 32.17 & 3.55 \\
  & DIUSum \shortcite{Xiao2024DIUSum:Summarization} & 42.23 & 19.83 & - & - & - \\
  & BERTAbs \shortcite{Xiao2024DIUSum:Summarization} & 41.85 & 19.40 & - & - & - \\
\midrule
\multirow{1}{*}{\textbf{Proposed Model}} 
  & DVP (ours)          & 44.20 & 20.77 & 74.03 & 31.68 & 3.55 \\
\bottomrule
\end{tabular}
\vspace{-0.3cm}
\caption{Results of published baseline models in existing studies.}
\label{tab:results}
\end{table*}

\begin{table*}[!t]
\vspace{-0.2cm}
\centering
\small 

\begin{tabular}{lccccccc}
\toprule
\textbf{System} & \textbf{R-1} & \textbf{R-2} & \textbf{BERTScore} & \textbf{IP} & \textbf{CLIPScore} & \textbf{MMAE} & \textbf{PCD}\\
\midrule
OneVision & 43.81 & 20.52 & 89.58  & 74.02 & 70.62 & 3.5447 & 32.66 \\
Vision Sampler & 44.06 & 20.78 & 89.53  & 74.01 & 70.54 &  3.5484 & 32.65 \\
DVP & \textbf{44.20} & 20.77 & 89.33  & \textbf{74.03} & 70.52 & \textbf{3.5521} & \textbf{32.81} \\
\bottomrule
\end{tabular}%
\caption{Model performance under multi-objective loss ($\mathcal{L}_{\text{MM}}$). PCD = Pairwise Cosine Distance (diversity).}\label{tab:ablation_multiobj}
\end{table*}

\subsection{Comparison with Existing Systems}
Besides the proposed DVP, we evaluate two variants that are simplifications of full DVP system: (i) \textbf{OneVision}, the base LLaVA-OneVision model; (ii) \textbf{Vision Sampler}, which omits the \gls{dvp} and directly feeds visual features into the gating mechanism; and (iii) \textbf{\gls{dvp}}, the full model shown in Figure~\ref{fig:framework}.

Since existing automatic benchmarks are primarily optimized for textual overlap, we report ROUGE to contextualize textual quality, while also reporting the standard MSMO-style visual metrics available from prior work. For our models we additionally report image-text alignment and within-set diversity proxies, noting that many published baselines do not provide these values, which limits direct comparisons on those axes.

 Table~\ref{tab:results} presents a comparison between our proposed model and a range of published multimodal summarization systems evaluated on the MSMO benchmark. These systems include early baselines like ATG, ATL, and HAN \cite{Zhu2018MSMO:Output}, as well as more recent models such as MOF \cite{Zhu2020MultimodalReference}, UniMS \cite{Zhang2022Unims:Distillation}, SITA \cite{Jiang2023ExploitingSummarization}, ViL-Sum \cite{Cui2024AlignSummarization}, and DIUSum \cite{Xiao2024DIUSum:Summarization}.

Among all systems, ViL-Sum achieves the highest ROUGE scores (44.29 for ROUGE-1 and 20.96 for ROUGE-2), indicating strong performance on textual overlap. SITA, on the other hand, achieves the highest Image Precision (IP) at 76.41, reflecting more accurate image selection with respect to ground-truth annotations.

Our model (\gls{dvp}), trained with a multi-objective loss, achieves ROUGE-1/44.20 and ROUGE-2/20.77, effectively matching ViL-Sum in text generation performance (a gap of only 0.09 and 0.19, respectively). In terms of image selection quality, \gls{dvp} attains IP/74.03, surpassing ViL-Sum (66.27) and approaching SITA's upper bound, while maintaining strong MaxSim and MMAE scores. This reflects \gls{dvp}'s ability to select images that are both relevant and semantically aligned with the summary.

These results are significant for two reasons. First, they confirm that our model competes effectively on traditional metrics despite incorporating deeper visual processing, even though such metrics are not well-suited for evaluating systems with stronger visual grounding. Second, they validate that \gls{dvp}'s image selection is not only quantitatively strong but also better aligned with the multimodal summarization objective, balancing informativeness, relevance, and non-redundancy more effectively than many existing approaches.

\subsection{Impact of Multi-Objective Training}

Table~\ref{tab:ablation_multiobj} shows that multi-objective training improves both textual and visual quality. \gls{dvp} recovers +0.39 ROUGE-1 and +0.19 ROUGE-2 compared to its single-objective counterpart (Table~\ref{tab:ablation_maskedlm}), while achieving the best visual diversity and IP among all evaluated models.

\subsection{Effect of Visual Processing Under MaskedLM}
\begin{table}[!tbp]
\vspace{-0.2cm}
\centering
\small 

\begin{tabular}{lccc}
\toprule
\textbf{System} & \textbf{R-1} & \textbf{R-2} & \textbf{BERTScore} \\
\midrule
OneVision & \textbf{44.26} & \textbf{20.86} & 89.12 \\
Vision Sampler & 43.89 & 20.61 & \textbf{89.54}\\
DVP & 43.81 & 20.58 & 89.50 \\
\bottomrule
\end{tabular}%
\vspace{-0.2cm}
\caption{Performance under MaskedLM training.}\label{tab:ablation_maskedlm}
\vspace{-0.4cm}
\end{table}

Table~\ref{tab:ablation_maskedlm} isolates the effect of introducing stronger visual processing under a MaskedLM training objective. Moving from OneVision to Vision Sampler and then to \gls{dvp} reduces ROUGE (from 44.26/20.86 for OneVision to 43.81/20.58 for \gls{dvp}), although BERTScore remains broadly comparable (89.12--89.54). This indicates that deeper visual processing does not automatically improve text-centric metrics and can even degrade n-gram overlap under a purely textual objective. This observation motivates our multi-objective formulation, which balances textual and visual objectives, partially recovers ROUGE, and improves visual selection metrics (Table~\ref{tab:results}).

\section{Analysis}
\subsection{Human Evaluation}

Automatic metrics alone may fail to capture fine-grained aspects of multimodal quality, particularly the alignment between textual and visual content. To address this limitation, we conduct a human evaluation of our \gls{dvp}-based system.

We sample 200 articles and collect three independent annotations per article via Amazon Mechanical Turk, resulting in 600 valid responses from 507 unique workers after filtering 2.1\% of submissions using attention checks. Annotators rate each example on a 5-point Likert scale across four dimensions: text quality, image relevance, image diversity, and overall multimodal quality.

As shown in Table~\ref{tab:human_eval_complete}, the system achieves consistently strong ratings across all dimensions, with most responses in the high-quality range ($\geq$4). Image relevance attains the highest average score, indicating strong alignment between images and text. Text quality and overall multimodal quality are also rated highly, suggesting that improved visual grounding does not compromise textual coherence.

Although image diversity scores are slightly lower, they remain positive, reflecting a reasonable balance between relevance and variety. Inter-annotator agreement is high for a subjective task, particularly under the within-one criterion, indicating consistent preferences among annotators. Overall, these findings provide complementary human evidence that the system produces coherent, visually grounded summaries.

\begin{table}[!tbp]
\centering
\small
\setlength{\tabcolsep}{4pt}
\begin{tabular}{lcccc}
\toprule
\textbf{Dimension} & \textbf{Mean (SD)} & \textbf{\% $\geq$4} & \textbf{Exact} & \textbf{W1} \\
\midrule
Text quality & 3.90 (0.69) & 80.1 & 49.0 & 90.0 \\
Image relevance & 4.04 (0.80) & 76.8 & 44.3 & 84.0 \\
Image diversity & 3.89 (0.83) & 73.2 & 43.0 & 82.2 \\
Overall quality & 4.00 (0.71) & 79.2 & 45.8 & 85.5 \\
\bottomrule
\end{tabular}
\caption{Human evaluation results. We report mean ratings and standard deviation (SD), the proportion of ratings $\geq$ 4, and inter-annotator agreement by reporting exact match and within-one agreement (W1) percentage.}
\label{tab:human_eval_complete}
\end{table}

\begin{figure*}[!tbp]
    \centering
    \includegraphics[width=0.8\textwidth]{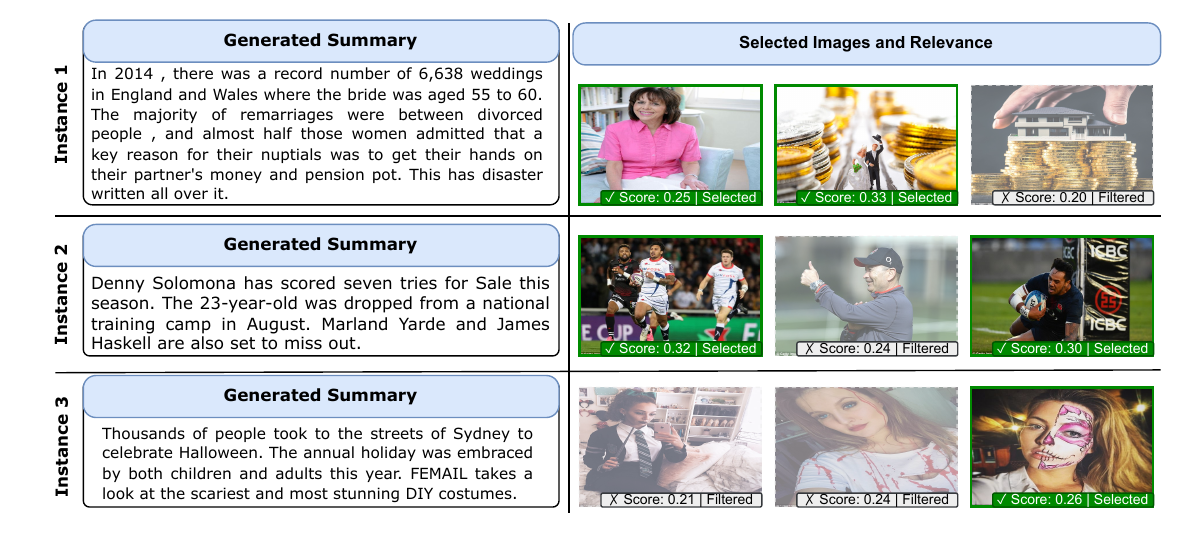}
    \caption{Examples from the test set illustrating image--summary relevance. For this experiment, a similarity threshold of 0.25 was used to filter out low-relevance images..}
    \label{fig:rel_vs_div}
\end{figure*}

\subsection{Relevance vs. Diversity Trade-off}\label{sec:relevance_filter}
As shown in Table~\ref{tab:relevance_filter}, diversity scores (measured via pairwise cosine distance) are highly sensitive to relevance filtering. Without filtering, semantically unrelated images inflate diversity scores artificially. After thresholding, the scores become more meaningful: \gls{dvp} maintains the highest mean and max diversity among filtered images, demonstrating its ability to select both relevant and non-redundant content.

\begin{table}[!tbp]
\centering
\small
\begin{tabular}{l c @{\hspace{1.5em}} c}
\toprule
System 
& w Rel. Filter
& w/o Rel. Filter \\
\cmidrule(lr){2-2} \cmidrule(lr){3-3}
& Mean / Max & Mean / Max \\
\midrule
OV             
& 32.66 / 36.53 
& 36.72 / 43.19 \\
Vision Sampler 
& 32.65 / 36.50 
& 36.73 / 43.22 \\
DVP            
& \textbf{32.81} / \textbf{36.71} 
& \textbf{36.77 }/ \textbf{43.27} \\
\bottomrule
\end{tabular}
\caption{Pairwise cosine distance among selected images (higher = more diverse).}
\label{tab:relevance_filter}
\end{table}

While embedding-based thresholds are effective for filtering irrelevant images, our qualitative analysis (Figure~\ref{fig:rel_vs_div}) reveals that some discarded images still provide useful contextual cues, such as related events or background details. Although not directly aligned with the summary content, these images can enrich reader understanding and serve a complementary role rather than being strictly irrelevant.

This raises an open question: How can models distinguish between irrelevant content and visually complementary information? Relying solely on hard similarity thresholds may discard valuable context. Future work should explore modeling complementarity alongside relevance and diversity, possibly incorporating user-centric or task-aware signals to better assess an image's supportive value.

\subsection{Qualitative Analysis of Visual Grounding}

\begin{figure*}[!t]
    \centering
    \includegraphics[width=0.8\textwidth]{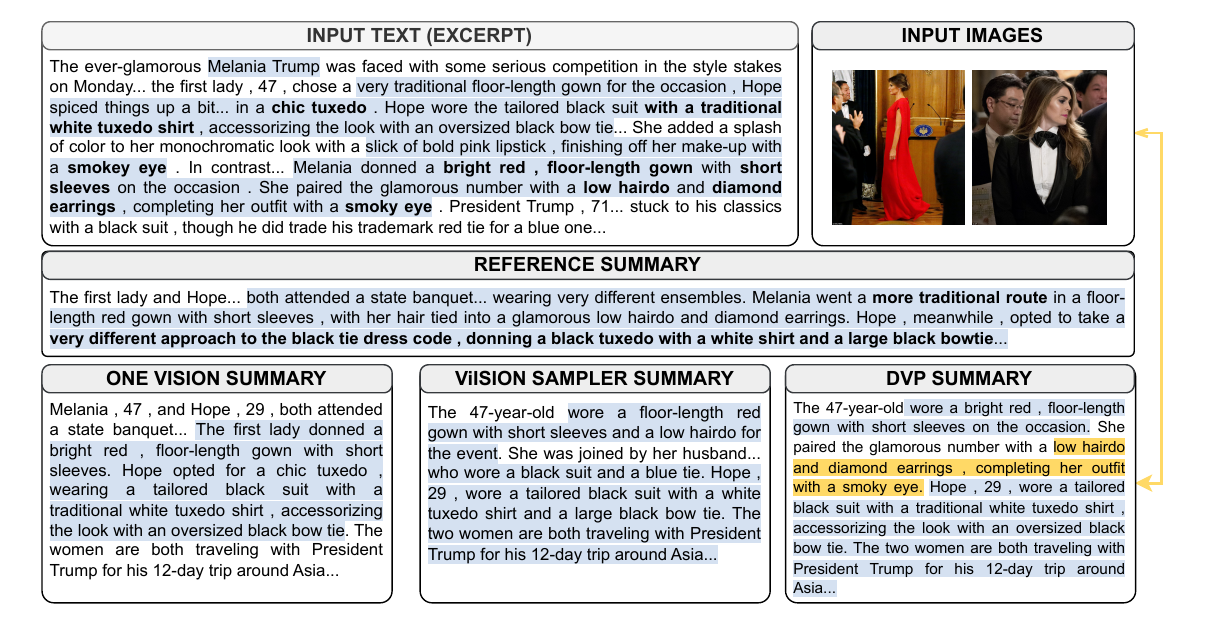}
    \vspace{-0.2cm}
    \caption{Comparison of summaries for a state banquet example. OV and Vision Sampler capture general descriptions (\textcolor{blue}{blue}), whereas \gls{dvp} includes fine-grained visual details (\textcolor{yellow}{yellow}), such as ``diamond earrings'' and ``smoky eye''.}
    \label{fig:output_1}
\end{figure*}

\begin{figure*}[!hbpt]
    \centering
    \includegraphics[width=0.8\textwidth]{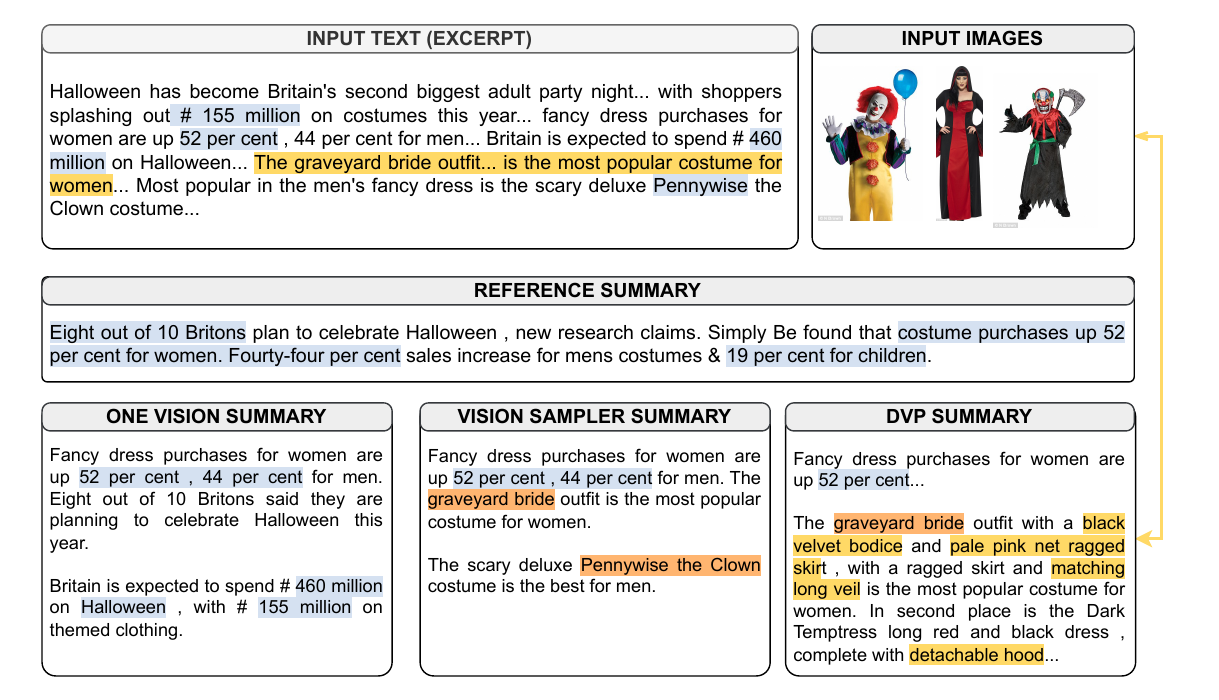}
    \caption{Comparison of summaries for Halloween costume trends. OV captures text-based statistics (\textcolor{blue}{blue}), Vision Sampler includes basic entities (\textcolor{orange}{orange}), and DVP adds fine-grained visual details (\textcolor{yellow}{yellow}) absent from the text.}
    \label{fig:output_2}
\end{figure*}

Our qualitative analysis on generated summaries (Figures~\ref{fig:output_1}--\ref{fig:output_2}) reveals a clear progression in model awareness. While baseline models (OV, Vision Sampler) exhibit a ``textual lead bias'' focused on procedural facts, \gls{dvp} leverages visual saliency as a priority filter. \gls{dvp} successfully extracts buried descriptive details, such as garment textures, specific accessories, and cultural markers, that are visually prominent in the source images but ignored by text-centric models. This demonstrates \gls{dvp}'s unique ability to re-prioritize information through cross-modal grounding, yielding summaries that better reflect the human viewing experience.

\section{Conclusion}
\label{sec:conclusion}

We introduced \texttt{SPeCTrA-Sum}, a multimodal summarization framework that elevates the role of visual evidence in text generation and performs principled, efficient image subset selection. Our approach incorporates a \glsentrylong{dvp} (\glsentryshort{dvp}) for layer-aligned visual conditioning and a lightweight \glsentrylong{vrp} (\glsentryshort{vrp}) distilled from a diversity-aware \gls{dpp} teacher. Together with multi-objective training, this design allows the model to produce not only fluent summaries but also visually grounded outputs that highlight the most informative elements of the input. Overall, \texttt{SPeCTrA-Sum} offers a cohesive and efficient solution to multimodal summarization, balancing informativeness, fluency, and visual complementarity.

\section*{Limitations}

Despite promising results, our study highlights several limitations in current evaluation practices and modeling assumptions for multimodal summarization.

First, standard automatic metrics like ROUGE remain poorly aligned with the goals of visually grounded generation. They measure n-gram overlap but not whether selected images meaningfully supplement the summary. While we report MSMO-style visual metrics and used additional proxies (e.g., CLIPScore, pairwise cosine distance), prior systems do not report these, limiting direct comparability.

Second, As discussed in Section~\ref{sec:relevance_filter}, diversity scores can be inflated by semantically irrelevant images, complicating interpretation unless standardized relevance filtering is applied.

Finally, while our evaluation focuses on automatic metrics, incorporating human judgments of visual-textual grounding, complementarity, and user-perceived utility would offer valuable additional insights. Developing such assessments and benchmarks could help more comprehensively capture multimodal coherence and further validate system performance.

\section*{Ethical Considerations}

Our work uses the MSMO dataset, derived from public news sources, which may reflect inherent social or cultural biases. While our model improves visual-textual alignment, it may inherit such biases and reflect subjective notions of image relevance. Future work should explore fairness-aware training and human evaluation to ensure broader and more responsible applicability.

\section*{Acknowledgments}
This research was undertaken with the assistance of resources from the National Computational Infrastructure (NCI Australia), an NCRIS enabled capability supported by the Australian Government.

\bibliography{references.bib}

\appendix

\section{Appendix}
\label{sec:appendix}

\subsection{Gated Cross-Attention Injection in Decoder}\label{sec:gated_cross_attention}
This appendix provides full implementation details for the gated cross-attention mechanism used to inject visual features into the decoder. We include the formulation of layer-wise cross-attention, gating behavior, and integration within the transformer block stack.

Let $\mathbf{h}^{(\ell)}\in\mathbb{R}^{T\times d}$ denote the LLM hidden states at layer $\ell$, and let the keys and values be derived from the depth-aligned visual states $\mathbf{v}^{(\ell)}$. The cross-attention output is computed as:

\begin{equation}
\begin{aligned}
\mathrm{XAttn}^{(\ell)}(\mathbf{h}^{(\ell)},\hat{\mathbf{v}}^{(\ell)})
&=
\mathrm{softmax}\!\left(
  \frac{%
    \begin{aligned}[t]
      Q^{(\ell)}\mathbf{h}^{(\ell)}\\[-1pt]
      \big(K^{(\ell)}\hat{\mathbf{v}}^{(\ell)}\big)^\top
    \end{aligned}
  }{\sqrt{d_k}}
\right) \\
\, V^{(\ell)}\hat{\mathbf{v}}^{(\ell)} \, ,
\end{aligned}
\label{eq:xattn}
\end{equation}

where $Q^{(\ell)}, K^{(\ell)}, V^{(\ell)}$ are learned projection matrices for queries, keys, and values respectively.

To modulate the contribution of the visual signal, we introduce a trainable gate $\alpha^{(\ell)} \in \mathbb{R}$, initialized close to zero. The residual connection is updated as:

\begin{equation}
\tilde{\mathbf{h}}^{(\ell)}=\mathbf{h}^{(\ell)}+\tanh\!\big(\alpha^{(\ell)}\big)\odot \mathrm{XAttn}^{(\ell)}(\mathbf{h}^{(\ell)},\hat{\mathbf{v}}^{(\ell)}),
\label{eq:gate}
\end{equation}

where $\odot$ denotes element-wise multiplication. Because $\tanh\!\big(\alpha^{(\ell)}\big) \approx 0$ at initialization, the model initially behaves like the original LLM, and learns over time how much visual information to integrate at each injection layer.

Between injection points, the LLM proceeds as usual with standard transformer blocks:

\begin{equation}
\mathbf{h}^{(\ell+1)}=\mathrm{LLMBlock}^{(\ell)}\!\left(\tilde{\mathbf{h}}^{(\ell)}\right).
\label{eq:llmblock}
\end{equation}

This design ensures that visual features are progressively and selectively fused, respecting both the depth and semantic abstraction level of the LLM's internal representations.

\subsection{DPP Teacher Kernel and Marginal Computation}\label{sec:dpp_derivation}
The \gls{dpp} kernel is constructed as:

\begin{equation}
    L = Q^{1/2} \kappa Q^{1/2} + \varepsilon I,
\end{equation}

where $Q = \mathrm{diag}(q_1, \ldots, q_K)$ encodes per-image relevance scores, $\kappa$ is an RBF-based similarity matrix capturing inter-image redundancy, and $\varepsilon I$ ensures numerical stability. To control the expected subset size, we define the marginal kernel $L(L + tI)^{-1}$, and solve for $t^*$ such that $\mathrm{Tr}(K(t^*)) = \mu$, the desired cardinality.
 
The resulting marginal inclusion probabilities are:
\begin{equation}
    \pi_i = \frac{\lambda_i}{\lambda_i + t^*},
\end{equation}

where $\lambda_i$ are the eigenvalues of $L$.

\subsection{Global Image-Text Alignment Objective}\label{sec:image_text_alignment_objective}
We encourage semantic consistency between the decoder's representation and the selected images using a SigLIP-inspired global contrastive loss.

Let $\mathrm{SigText}$ and $\mathrm{SigImage}$ be frozen encoders that map inputs into  $\ell_2$-normalized vectors in $\mathbb{R}^{d_p}$. For an article $a$, with reference summary $Y_a$ and selected images $\{I^a_{i_k}\}_{k=1}^{K}$, the teacher embeddings are defined as:

\begin{equation}
\begin{aligned}
t_{\text{sig}}(a)
&= \mathrm{norm}\!\big(\mathrm{SigText}(Y_a)\big), \\[4pt]
i_k^a
&= \mathrm{norm}\!\big(\mathrm{SigImage}(I^a_{i_k})\big), \\[4pt]
v_{\text{sig}}(a)
&= \mathrm{norm}\!\Big(\tfrac{1}{K}\sum_{k=1}^{K} i_k^a\Big).
\end{aligned}  
\end{equation}

Here, $v_{\text{sig}}(a)$ is the mean pooled image embedding for article $a$, encoding the visual summary.

On the student side, let $H^{\text{dec}}\in\mathbb{R}^{T\times d}$ denote the decoder hidden states under teacher forcing. We compute the student's pooled representation via:

\begin{equation*}
    s = \mathrm{meanpool}(H^{\text{dec}}),\; t_{\text{stu}} = \mathrm{norm}\!\big(g_{\text{text}}(s)\big)
\end{equation*}

where $g_{\text{text}}:\mathbb{R}^{d}\!\to\!\mathbb{R}^{d_p}$ is a learnable projection module that maps decoder features to the alignment space.

We define a logistic similarity function:

\begin{equation}
    z(x,y)=\tfrac{1}{\tau}\,x^\top y
\end{equation}

with temperature parameter $\tau>0$, and draw a set of negatives $\mathcal{S}$ from other articles in the batch or dataset.

The image-text alignment loss is then formulated in a SigLIP-style contrastive objective:

\begin{equation}
\begin{aligned}
\mathcal{L}_{\text{align}}
&=
-\log \sigma\!\big(z(t_{\text{stu}},\,v_{\text{sig}}(a))\big)
\\[3pt]
&\quad
-\frac{1}{|\mathcal{S}|}\sum_{j\in\mathcal{S}}\log\!\big(1-\sigma(z(t_{\text{stu}},\,v_{\text{sig}}(j)))\big)
\\[3pt]
&\quad
-\frac{1}{|\mathcal{S}|}\sum_{j\in\mathcal{S}}\log\!\big(1-\sigma(z(t_{\text{sig}}(j),\,v_{\text{sig}}(a)))\big) .
\end{aligned}
\label{eq:global_loss}
\end{equation}

where $\sigma$ is the sigmoid function. The first term maximizes similarity between the student text and corresponding visual summary. The second penalizes similarity between the student text and mismatched visual contexts. The third penalizes mismatched teacher text--image pairs to regularize the frozen teacher space.

This alignment objective encourages the student decoder to produce summaries that are semantically grounded in the selected images, while also aligning to a frozen, high-quality multimodal embedding space.

\subsection{DDP-Based Image Selection and Distillation}\label{sec:dpp_image_selection}
This appendix provides full details on the construction of the \gls{dpp} teacher used to supervise the \gls{vrp}, including the derivation of marginal probabilities and the training objective used for distillation.

\subsubsection{DPP Teacher Construction}
Let ${I^a_i}_{i=1}^{K}$ be the candidate images for a document or article $a$, with reference summary $Y_a$. The \gls{dpp} teacher produces a probability distribution over subsets of images that favors both high relevance to the text and inter-image diversity.

We use a frozen SigLIP model to extract $\ell_2$-normalized embeddings for the summary and each image:

\begin{equation}
\begin{aligned}
e_{\text{text}}
&= \mathrm{norm}\!\big(\mathrm{SigText}(Y_a)\big), \\[3pt]
e_i
&= \mathrm{norm}\!\big(\mathrm{SigImage}(I^a_i)\big).
\end{aligned}
\label{eq:embeddings}
\end{equation}

The relevance score for image $i$ is computed as the inner product $r_i=e_{\text{text}}^\top e_i$ , and similarity between image $i$ and $j$ as $s_{ij}=e_i^\top e_j$.

We compute per-image quality scores as:
\begin{equation*}
    q_i=\exp(\gamma r_i)
\end{equation*}

where $\gamma > 0$ is a scaling factor.

To model diversity, we define an RBF kernel on the cosine distances between image embeddings:
\begin{equation*}
    D^2_{ij}=2\,(1-s_{ij}),\,
    \kappa_{ij}=\exp\!\Big(-\tfrac{D^2_{ij}}{2\sigma^2}\Big),\, \kappa_{ii}=1,
\end{equation*}

\paragraph{DPP Kernel:} We construct a quality-diversity kernel $L \in \mathbb{R}^{K \times K}$ as:
\begin{equation}\label{eq:L_kernel}
    L \;=\; Q^{1/2}\,\kappa\,Q^{1/2} \;+\; \varepsilon I,
    \,
    Q=\mathrm{diag}(q_1,\dots,q_K),
\end{equation}

where $\varepsilon > 0$ is a small constant for numerical stability. This kernel defines a \gls{dpp} over image subsets.

\subsubsection{Marginal Inclusion Probabilities}
To control the expected number of selected images $\mu$, we define the marginal kernel:
\begin{equation}
\label{eq:K_marginal}
K(t) \;=\; L\,(L+tI)^{-1},\quad t\ge0,
\end{equation}

and solve for $t^*$ such that:
\begin{equation*}
    \mathrm{Tr}\,K(t^*)=\mu
\end{equation*}

Using eigendecomposition $L=U\,\mathrm{diag}(\lambda_\ell)\,U^\top$, we compute:

\begin{equation}
\begin{aligned}
\mathrm{Tr}\,K(t)
&= \sum_{\ell}\frac{\lambda_\ell}{\lambda_\ell + t}, \\[4pt]
K(t)
&= U\,\mathrm{diag}\!\Big(\tfrac{\lambda_\ell}{\lambda_\ell + t}\Big)\,U^\top .
\end{aligned}
\label{eq:kernel_trace}
\end{equation}

The final soft marginal inclusion probabilities used for distillation are:
\begin{equation}
\label{eq:teacher_marginals}
\pi_i \;=\; [K(t^*)]_{ii}
\;\in\;[0,1],
\quad
\sum_{i=1}^{K}\pi_i \;\approx\; \mu,
\end{equation}

These soft labels encoder: (1) Relevance (via $q_i$), (2) Diversity (via $\kappa$), and (3) Cardinality control (via $\mu$ constraint).

\subsubsection{VRP Student Objective}
Let $z_i = f_{\text{VRP}}(v_i)$ denote the scalar logit predicted by the \gls{vrp} for image $i$, and $p_i = \sigma(z_i)$ the corresponding selection probability.

We train the \gls{vrp} to match the \gls{dpp} teacher marginals using a stable logit-level binary cross-entropy, plus a soft constraint on expected cardinality:
\begin{equation}
\begin{aligned}
\mathcal{L}_{\text{VRP}}
&=
\underbrace{\tfrac{1}{K}\sum_{i=1}^{K}
   \Big(\mathrm{softplus}(z_i) - \pi_i z_i\Big)}_{\text{probability matching}}
\\[4pt]
&\quad
+\;
\underbrace{\alpha\,\Big(\sum_{i=1}^{K} p_i - \mu\Big)^2}_{\text{cardinality regularization}} .
\end{aligned}
\label{eq:vrp_loss}
\end{equation}

where $\alpha \ge 0$ controls the strength of the regularization.

\noindent This distillation approach transfers the inductive bias of \glspl{dpp} into a lightweight, text-agnostic image selector. It enables:
\begin{itemize}
    \item $\mathcal{O}(K)$ inference (vs. $\mathcal{O}(K^3)$ for \gls{dpp} sampling),
    \item Soft, diverse image selection at test time,
    \item And stable training via smooth supervision.
\end{itemize}

This loss complements our summarization and alignment objectives, resulting in coherent summaries grounded in relevant and diverse visual evidence.

\subsection{Technical Implementation Details}\label{app:technical_details}
\subsubsection{Technical Details}
We used LLaVA-OneVision with Qwen2-7B variant as our base model. On top of the base model, we introduced all the modules discussed in Section~\ref{sec:proposed_method}. Key training hyperparameters are listed in Table~\ref{tab:training_parameters}.

\begin{table}[!htbp]
\centering

\begin{tabular}{ll}
\toprule
Hyperparameter & Value \\
\midrule
Optimizer & Adafactor \\
Mixed Precision & bfloat16 \\
Precision & bf16 \\
Max input length & 2048 \\
Max images & 5 \\
\bottomrule
\end{tabular}%
\caption{Key training hyperparameters.}\label{tab:training_parameters}
\end{table}

We use a two-stage training procedure guided by the validation loss focusing on Architecture (Vision Sampler/\gls{dvp} capacity and the set of gated decoder layers). Then LoRA fine-tuning parameters were optimized using the best architecture from stage 1. Search space and relevant hyperparameter are listed in Table~\ref{tab:group_hyperparameter}.

Regarding image selection using \gls{vrp} and \gls{dpp}, the hyperparameter details and their interpretation is given in Table~\ref{tab:vrp_paraemeters}. Moreover, all experiments were conducted using a single NVIDIA A100 GPU with 80 GB memory. For memory efficiency, 4-bit QLoRA-style quantization technique is used.

\begin{table}[!htbp]
\centering
\small 

\begin{tabular}{l p{4cm} c}
\toprule
H.parameter & Interpretation & Value \\
\midrule
$K$ & max selected images per example & 3 \\
$\sigma$ & similarity kernel bandwidth (RBF) & 0.8 \\
$\gamma$ & relevance scaling for $q_i$ (quality weighting) & $2.0$ \\
$\mu$ & target/expected subset size (cardinality prior) & $3.0$ \\
$\alpha$ & strength of subset-size regularization & $0.3$ \\
$\epsilon$ & numerical stability term for kernel matrix & $1\times 10^{-5}$  \\
$\gamma$ & relevance scaling for $q_i$ (quality weighting) & $2.0$ \\
\bottomrule
\end{tabular}%
\caption{VRP-related hyperparameters.}\label{tab:vrp_paraemeters}
\end{table}

\subsubsection{Dataset Details}
Sourced from the Daily Mail website, MSMO pairs each news article with all in-page images, and constructs reference summaries by combining article headlines. Human annotators also mark the images deemed most representative of the story. On average, each article contains approximately 650 words and nine images, offering a realistic and visually rich setting for multimodal summarization. The dataset provides both textual references (for ROUGE-style evaluation) and image relevance labels. Owing to its scale, permissive licensing, and established leaderboard, MSMO has become the de-facto benchmark---often described as the ``MNIST of multimodal summarization''.

\begin{table*}[!htbp]
\centering
\begin{tabular}{l l l}
\toprule
Group & Hyperparameter & Search Space and Optimized Values \\
\midrule
\multirow{3}{*}{Vision Sampler}
& Number of latents & \text{[\textbf{32}, 64]} \\
& Depth & {[2, \textbf{4}]} \\
& FF multiplier & {[2, \textbf{4}]} \\ [6pt]
DVP & Number of layers & \text{[16, 20, \textbf{24}]} \\ [6pt]
\multirow{4}{*}{Gated Connections}
& \multirow{4}{*}{Layer indices}
& \text{[8, 16] with 16 DVP layers} \\
& &
\text{[8, 16] with 20 DVP layers}  \\
& &
\textbf{[8, 16, 24] with 24 DVP layers}  \\
& &
\text{[4, 8, ... , 24] with 24 DVP layers}  \\ [6pt]

\multirow{2}{*}{LoRA}
& Rank $r$ & \text{[16, \textbf{32}, 64]} \\
& $\alpha $ & \text{[64, \textbf{128}]} \\
\bottomrule
\end{tabular}
\caption{Search space for each hyperparameter group with optimal values highlighted in \textbf{bold}.}
\label{tab:group_hyperparameter}
\end{table*}

\subsection{Computational Overhead Analysis}\label{sec:overhead_analysis}
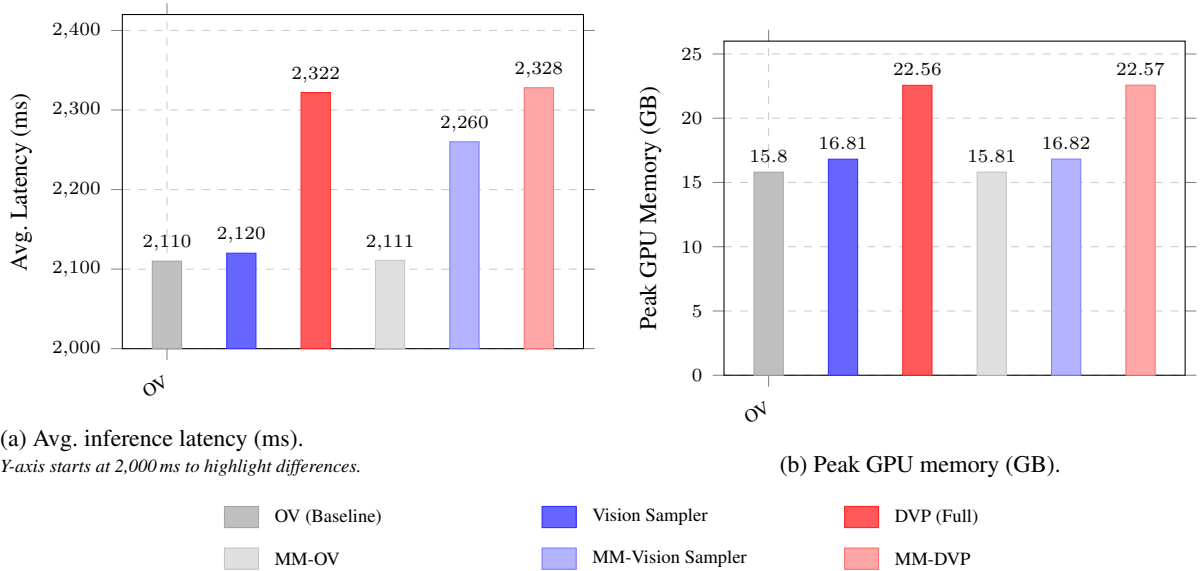
\begin{figure*}[ht]
  \centering

  \begin{subfigure}[b]{0.48\textwidth}
    \begin{tikzpicture}
      \begin{axis}[
        ybar,
        bar width        = 11pt,
        width            = \linewidth,
        height           = 6cm,
        enlarge x limits = 0.12,
        ymin = 2000, ymax = 2420,
        ylabel           = {Avg.\ Latency (ms)},
        ylabel style     = {font=\small},
        xtick            = data,
        xticklabels      = {OV, Vis.\ Sam., DVP, MM-OV, MM-VS, MM-DVP},
        xticklabel style = {font=\scriptsize, rotate=30, anchor=north east},
        ytick            = {2000, 2100, 2200, 2300, 2400},
        yticklabel style = {font=\scriptsize},
        tick align       = outside,
        axis line style  = {-},
        grid             = major,
        grid style       = {dashed, gray!40},
        nodes near coords,
        nodes near coords align             = {above},
        every node near coord/.append style = {font=\scriptsize},
      ]
        \addplot[fill=gray!50,  draw=gray!70,  bar shift=0pt]  coordinates {(1, 2110)};
        \addplot[fill=blue!60,  draw=blue!80,  bar shift=0pt]  coordinates {(2, 2120)};
        \addplot[fill=red!65,   draw=red!85,   bar shift=0pt]  coordinates {(3, 2322)};
        \addplot[fill=gray!25,  draw=gray!45,  bar shift=0pt]  coordinates {(4, 2111)};
        \addplot[fill=blue!30,  draw=blue!50,  bar shift=0pt]  coordinates {(5, 2260)};
        \addplot[fill=red!35,   draw=red!55,   bar shift=0pt]  coordinates {(6, 2328)};
      \end{axis}
    \end{tikzpicture}
    \caption{Avg.\ inference latency (ms).\newline
             {\scriptsize\textit{Y-axis starts at 2,000\,ms to highlight differences.}}}
    \label{fig:overhead_latency}
  \end{subfigure}
  \hfill
  \begin{subfigure}[b]{0.48\textwidth}
    \begin{tikzpicture}
      \begin{axis}[
        ybar,
        bar width        = 11pt,
        width            = \linewidth,
        height           = 6cm,
        enlarge x limits = 0.12,
        ymin = 0, ymax   = 26,
        ylabel           = {Peak GPU Memory (GB)},
        ylabel style     = {font=\small},
        xtick            = data,
        xticklabels      = {OV, Vis.\ Sam., DVP, MM-OV, MM-VS, MM-DVP},
        xticklabel style = {font=\scriptsize, rotate=30, anchor=north east},
        ytick            = {0, 5, 10, 15, 20, 25},
        yticklabel style = {font=\scriptsize},
        tick align       = outside,
        axis line style  = {-},
        grid             = major,
        grid style       = {dashed, gray!40},
        nodes near coords,
        nodes near coords align             = {above},
        every node near coord/.append style = {
          font=\scriptsize,
          /pgf/number format/.cd, fixed, precision=2},
      ]
        \addplot[fill=gray!50,  draw=gray!70,  bar shift=0pt]  coordinates {(1, 15.80)};
        \addplot[fill=blue!60,  draw=blue!80,  bar shift=0pt]  coordinates {(2, 16.81)};
        \addplot[fill=red!65,   draw=red!85,   bar shift=0pt]  coordinates {(3, 22.56)};
        \addplot[fill=gray!25,  draw=gray!45,  bar shift=0pt]  coordinates {(4, 15.81)};
        \addplot[fill=blue!30,  draw=blue!50,  bar shift=0pt]  coordinates {(5, 16.82)};
        \addplot[fill=red!35,   draw=red!55,   bar shift=0pt]  coordinates {(6, 22.57)};
      \end{axis}
    \end{tikzpicture}
    \caption{Peak GPU memory (GB).}
    \label{fig:overhead_memory}
  \end{subfigure}

  \vspace{6pt}
  \begin{tikzpicture}[font=\scriptsize]
    \def\sw{0.45cm}\def\sh{0.28cm}\def\vsep{0.08cm}
    \fill[gray!50]  (0,0)     rectangle (\sw,\sh); \draw[gray!70]  (0,0)     rectangle (\sw,\sh);
    \node[right] at (\sw+\vsep, 0.5*\sh) {OV (Baseline)};
    \fill[blue!60]  (4.2cm,0) rectangle (4.2cm+\sw,\sh); \draw[blue!80]  (4.2cm,0) rectangle (4.2cm+\sw,\sh);
    \node[right] at (4.2cm+\sw+\vsep, 0.5*\sh) {Vision Sampler};
    \fill[red!65]   (8.2cm,0) rectangle (8.2cm+\sw,\sh); \draw[red!85]   (8.2cm,0) rectangle (8.2cm+\sw,\sh);
    \node[right] at (8.2cm+\sw+\vsep, 0.5*\sh) {DVP (Full)};
    \def\yy{-0.55cm}
    \fill[gray!25]  (0,\yy)     rectangle (\sw,\yy+\sh); \draw[gray!45]  (0,\yy)     rectangle (\sw,\yy+\sh);
    \node[right] at (\sw+\vsep, \yy+0.5*\sh) {MM-OV};
    \fill[blue!30]  (4.2cm,\yy) rectangle (4.2cm+\sw,\yy+\sh); \draw[blue!50]  (4.2cm,\yy) rectangle (4.2cm+\sw,\yy+\sh);
    \node[right] at (4.2cm+\sw+\vsep, \yy+0.5*\sh) {MM-Vision Sampler};
    \fill[red!35]   (8.2cm,\yy) rectangle (8.2cm+\sw,\yy+\sh); \draw[red!55]   (8.2cm,\yy) rectangle (8.2cm+\sw,\yy+\sh);
    \node[right] at (8.2cm+\sw+\vsep, \yy+0.5*\sh) {MM-DVP};
  \end{tikzpicture}

  \caption{Absolute inference cost of all six model variants
           (exact figures in Table~\ref{tab:inference_speed_memory}).
           \textbf{Left:} average latency per sample (ms); y-axis starts at
           2{,}000\,ms to accentuate differences --- actual values are printed
           above each bar.
           \textbf{Right:} peak GPU memory (GB).
           Darker shades = base architecture; lighter shades = corresponding
           MM-loss variant (inference-identical to their base counterpart).}
  \label{fig:overhead}
\end{figure*}

We provide a detailed computational overhead analysis comparing \gls{dvp} against the simple concatenation baseline (OV) across all six model variants. All measurements were performed on a single NVIDIA A100-SXM4-80GB GPU using 100 test samples with 10 warm-up iterations, generating up to 256 tokens per sample under identical hardware and software conditions. Table~\ref{tab:inference_speed_memory} reports exact figures for all variants;
Figure~\ref{fig:overhead} visualises the absolute latency and memory cost side by side.

\begin{table*}[ht]
  \centering

  \resizebox{\textwidth}{!}{%
  \begin{tabular}{llcccc}
    \toprule
    Variant        & Architecture           & Avg.\ Latency (ms)  & Latency Overhead
                   & Peak GPU Mem.\ (GB)   & Memory Overhead \\
    \midrule
    OV (Baseline)    & Concatenation               & ${\sim}2{,}110$     & ---       & 15.80 & ---      \\
    Vision Sampler   & Perceiver Only              & $2{,}120$           & $+0.5\%$  & 16.81 & $+6.4\%$ \\
    DVP              & DVP (Full)                  & $2{,}322$           & $+10.0\%$ & 22.56 & $+42.8\%$\\
    MM-OV            & Concat + MM Loss            & $2{,}111$           & $+0.0\%$  & 15.81 & $+0.1\%$ \\
    MM-Vision Sampler& Perceiver + MM Loss         & $2{,}260$           & $+7.1\%$  & 16.82 & $+6.5\%$ \\
    MM-DVP           & DVP + MM Loss               & $2{,}328$           & $+10.3\%$ & 22.57 & $+42.8\%$\\
    \bottomrule
  \end{tabular}%
  }
  \caption{Inference speed and peak GPU memory across all six model variants.
           Measurements on a single NVIDIA A100-SXM4-80GB GPU (100 test
           samples, 10 warm-up iterations, up to 256 generated tokens).}
  \label{tab:inference_speed_memory}
\end{table*}

\end{document}